\documentclass[sigconf]{acmart}
\AtBeginDocument{%
  \providecommand\BibTeX{{%
    \normalfont B\kern-0.5em{\scshape i\kern-0.25em b}\kern-0.8em\TeX}}}

\copyrightyear{2021} 
\acmYear{2021} 
\setcopyright{acmcopyright}\acmConference[MM '21]{Proceedings of the 29th ACM International Conference on Multimedia}{October 20--24, 2021}{Virtual Event, China}
\acmBooktitle{Proceedings of the 29th ACM International Conference on Multimedia (MM '21), October 20--24, 2021, Virtual Event, China}
\acmPrice{15.00}
\acmDOI{10.1145/3474085.3475309}
\acmISBN{978-1-4503-8651-7/21/10}

\acmSubmissionID{813}

\usepackage{amssymb}
\usepackage{url}

\usepackage{mmstyles}

\newcommand{\method}{VoteHMR}

\begin{document}
\fancyhead{}

\title{VoteHMR: Occlusion-Aware Voting Network for Robust 3D Human Mesh Recovery from Partial Point Clouds}



\author{Guanze Liu}
\affiliation{
    \institution{College of Software, Beihang University}
    \city{Beijing}
    \country{China}
}
\email{sy1921112@buaa.edu.cn}

\author{Yu Rong}
\affiliation{
    \institution{The Chinese University of Hong Kong}
    \city{Hong Kong SAR}
    \country{China}
}
\email{rongyu9124@gmail.com}

\author{Lu Sheng}
\authornote{Lu Sheng is the corresponding author.}
\affiliation{
        \institution{College of Software, Beihang University}
    \city{Beijing}
    \country{China}
}
\email{lsheng@buaa.edu.cn}







\begin{abstract}
  3D human mesh recovery from point clouds is essential for various tasks, including AR/VR and human behavior understanding. Previous works in this field either require high-quality 3D human scans or sequential point clouds, which cannot be easily applied to low-quality 3D scans captured by consumer-level depth sensors. 
  In this paper, we make the first attempt to reconstruct reliable 3D human shapes from single-frame partial point clouds.
  To achieve this, we propose an end-to-end learnable method, named \method. 
  The core of \method~is a novel occlusion-aware voting network that can first reliably produce visible joint-level features from the input partial point clouds, and then complete the joint-level features through the kinematic tree of the human skeleton.
  Compared with holistic features used by previous works, the joint-level features can not only effectively encode the human geometry information but also be robust to noisy inputs with self-occlusions and missing areas. 
  By exploiting the rich complementary clues from the joint-level features and global features from the input point clouds, the proposed method encourages reliable and disentangled parameter predictions for statistical 3D human models, such as SMPL.
  The proposed method achieves state-of-the-art performances on two large-scale datasets, namely SURREAL and DFAUST. Furthermore, \method~also demonstrates superior generalization ability on real-world datasets, such as Berkeley MHAD.
\end{abstract}

\ccsdesc[500]{Computing methodologies~Point-based models}

\ccsdesc[500]{Computing methodologies~Shape representations}

\ccsdesc[500]{Computing methodologies~Shape inference}

\ccsdesc[500]{Computing methodologies~Reconstruction}

\keywords{3D human shape reconstruction, occlusion handling, hough voting in point clouds}


\maketitle

\section{Introduction}

Recovering 3D human shapes is an appealing yet challenging task in the research community of 3D computer vision and multimedia.
It can significantly benefit various downstream applications including AR/VR~\cite{wang2021scene}, robotics~\cite{Peng2018VDB,2021-TOG-AMP} and \emph{etc}.
This task aims at predicting the personalized 3D human models of the captured targets into non-parametric meshes~\cite{kolotouros2019convolutional} or parametric descriptions~\cite{kanazawa2018end, guler2019holopose, jiang2019skeleton, zhang2020learning} of a statistical body model~\cite{anguelov2005scape, loper2015smpl, osman2020star}.
%
In this paper, we focus on human mesh recovery from the point clouds captured by consumer-level depth sensors, such as Kinect and ToF cameras.
%
The irregular, sparse and orderless nature of this special 3D input brings additional challenges to designing a robust point-based reconstruction system, which cannot be achieved by simply adopting the ideas of recent success in RGB-based approaches~\cite{kanazawa2018end, kolotouros2019convolutional}.

\begin{figure}[t]
\centering
\includegraphics[width=\linewidth]{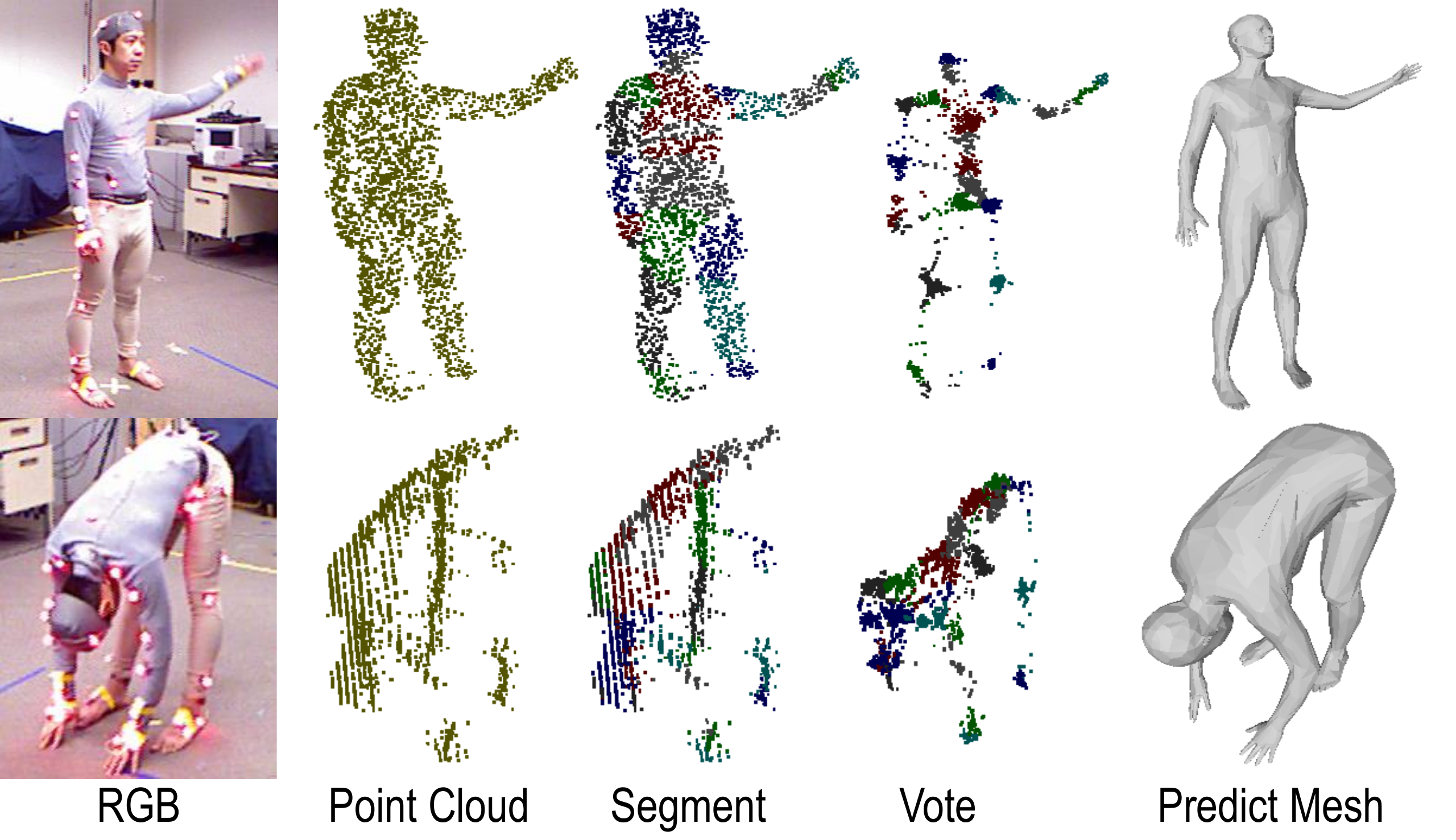}
\vspace{-5mm}
\caption{Human mesh recovery from a single frame of partial point clouds usually suffers from occluded human body parts and noises. In our \method, we adopt a robust voting network to reliably cluster visible joint-level features, showing impressive reconstruction onto challenging samples in the Berkerly MHAD dataset~\cite{ofli2013berkeley}.}
\label{fig:intro}
\end{figure}

Although promising reconstruction results have been demonstrated by early attempts that leverage a complete point cloud of a human scan~\cite{jiang2019skeleton} or a sequence of point clouds~\cite{wang2020sequential}, existing approaches are not yet able to reliably cope with a single frame of partial point clouds that are affected with severe self-occlusions and noises. 
Many applications (such as on mobile devices) may not afford to generate a complete scan of the captured targets or store a sequence of point clouds before the process of 3D human shape reconstruction.
Thus, it is meaningful to study robust 3D human mesh recovery from a single frame of partial point clouds. 
To achieve this, there are several key challenges that need to be addressed: (1) Coping with complex self-occlusions and missing parts for robust pose estimation, (2) Sustaining reasonable shape reconstruction without the interference of noises.

In this paper, we propose an occlusion-aware voting network for robust human shape reconstruction from a single frame of partial point clouds, named \method, that tackles the aforementioned challenges using an end-to-end learnable framework.
Our \method~ is designed to regress the parameters for statistical human body models, such as SMPL~\cite{loper2015smpl}, rather than directly learning the regression of the 3D coordinates of a high-dimensional human mesh.
This design can eliminate the distortions on the generated meshes when interfered with by the flaws of the input point clouds.
Moreover, inspired that the Hough voting~\cite{hough1959machine} has been successfully applied for object detection on the point clouds~\cite{votenet}, 
we adopt a similar mechanism to gather sparse points from visible human parts for discovering observed skeleton joints and their shape representations.
%
These occlusion-aware joint-level features are then completed by leveraging the kinematic tree of human skeletons.
Compared with holistic features~\cite{kocabas2020vibe, kanazawa2018end}, these joint-level features not only explicitly encode the human poses but also implicitly extract the local geometric contexts in each body part, thus benefit for comprehensive but disentangled predictions of the shape and pose of the captured targets.

To be specific, in our ~\method, we first propose a partial voting generation and clustering module to gather enough reliable points to discover the joint-level shape representations around the visible 3D skeleton joints.
After that, an occlusion-aware joint completion module is leveraged to complete the joint-level representations throughout the kinematic tree.
Thereafter, we propose two parallel global parameter regression and local parameter regression modules to disentangle the prediction of shape and pose parameters for the SMPL model.
The whole network is fully supervised by synthetic datasets such as SURREAL~\cite{varol2017learning} and DFAUST~\cite{bogo2017dynamic}, where the ground-truth (GT) human meshes, 3D coordinates of skeleton joints, and body part segmentation labels can be generated for each input point cloud.
It is worth noting that our method can be readily tested on real point clouds captured by consumer-level depth sensors with superior reconstruction performances than the existing methods~\cite{groueix20183d,wang2020sequential,jiang2019skeleton}.
To demonstrate this, we evaluate the trained models on Berkeley MHAD dataset~\cite{ofli2013berkeley}. Several results are shown in Figure~\ref{fig:intro}.
It is revealed that \method~ can reliably vote the visible joints with their associated body part segments, and then recover the human meshes with diverse poses from raw point clouds captured by Kinect. 
Note that \method~is just tested on this dataset, which validates that the proposed method is robust and generalizable to real-world scenarios.

In summary, the contributions of this paper are threefold: (1) The first method, named as \method, that successfully apply Hough voting to achieve parametric human shape reconstruction from a single frame of partial point clouds. (2) Certain robustness over complex occlusions, noises, and sparsity that frequently occur in the point clouds. (3) The state-of-the-art human mesh recovery performances on two benchmark datasets, SURREAL~\cite{varol2017learning} and DFAUST~\cite{bogo2017dynamic}, with additional nice generalization towards real-world datasets such as Berkeley MHAD~\cite{ofli2013berkeley}.

\section{Related Works}

\subsection{Human Mesh Recovery from 2D Images}

Human mesh recovery has been extensively studied from RGB images, either by template-based approaches to fit the 2D annotations of the RGB images~\cite{kanazawa2019learning, pavlakos2017coarse, pavlakos2018learning, varol2018bodynet, verma2018feastnet, pavlakos2019expressive, kolotouros2019learning, rong2020frankmocap, rong2020chasing, wang2020motion}, such as keypoints~\cite{smplify}, silhouettes~\cite{pavlakos2018learning}, and dense annotations~\cite{guler2018densepose, rong2019delving}, or template-less approaches~\cite{pavlakos2017coarse, pavlakos2018learning, saito2019pifu, zheng2019deephuman, zhu2019detailed} that directly regress the 3D coordinates of vertices of the human meshes.
The template-based methods often make use of human priors such as skeletons or parametric models like SCAPE~\cite{anguelov2005scape} and SMPL~\cite{loper2015smpl}, thus can alleviate the twisted and rugged human shapes that may occur by template-less methods.
Recent works also try to process depth data~\cite{pavlakos2017coarse, wang2020sequential, jiang2019skeleton, bogo2015detailed, groueix20183d, wei2016dense} for human mesh recovery, in which the existing methods can also be divided into template-based~\cite{ye2012performance, ye2014real, bogo2015detailed} and template-less methods~\cite{newcombe2015dynamicfusion, innmann2016volumedeform, dou20153d, dou2016fusion4d}, in which the template-less methods create the 3D human meshes without any prior knowledge about the
body shape and usually volumetrically fuse all captured
depth maps to reconstruct 3D models.
Some works~\cite{zheng2018hybridfusion, yu2017bodyfusion, yu2018doublefusion} try to combine template-priors and template-less methods for more robust mesh recovery with large human motions.

\begin{figure*}
\centering
\includegraphics[width=\linewidth]{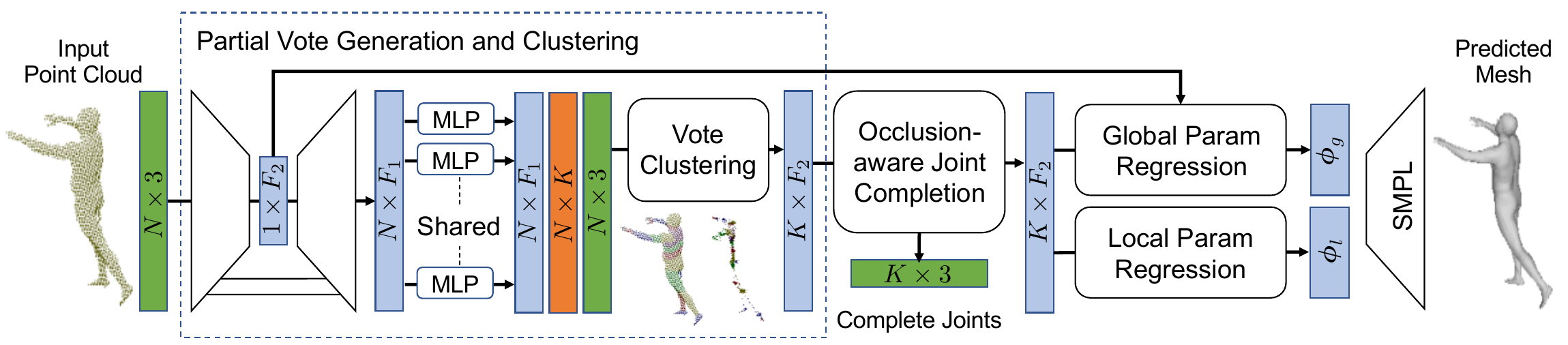}
\vspace{-5mm}
\caption{Overview of the proposed VoteHMR method. Best viewed on screen.}
\label{fig:framework}
\end{figure*}

\subsection{Human Mesh Recovery from Point Clouds}

Recent works have also tried to recover 3D human meshes from point clouds.
These works often utilize a pointnet~\cite{qi2017pointnet} or pointnet++~\cite{qi2017pointnet++} backbone to extract features from orderless point clouds. 
Jiang~\etal~\cite{jiang2019skeleton} propose a skeleton-aware network to predict human mesh from complete human point cloud scans but failed to generalize well to partial human scans.
Groueix~\etal~\cite{groueix20183d} deformed a template model to a given point cloud but often produce rugged human mesh.
Wang \etal~\cite{wang2020sequential} used a coarse-to-fine spatial-temporal mesh attention convolution module to predict human meshes from sequences of point clouds.
In their framework, temporal dependencies are vital to rectify the flaws caused by partial scans.
Few works have been proposed to tackle occlusions from a single frame of point clouds.
Since existing methods are usually trained on synthetic datasets like~\cite{varol2017learning, bogo2017dynamic}, these methods~\cite{groueix20183d, wang2020sequential, jiang2019skeleton} usually fail to generalize well to real scenarios and require a weakly supervised fine-tuning scheme. 
Unlike these methods, We provide an occlusion-aware voting-based human mesh recovery network that proves to be robust against noise and occlusions from a single frame of partial point clouds. We train our~\method~ on synthetic datasets~\cite{varol2017learning, bogo2017dynamic}, but it can generalize well to real-world scenarios.

\subsection{Voting Methods in Deep Learning}

Traditional Hough voting~\cite{hough1959machine} translates the problem of detecting patterns to detecting peaks in the parametric space.
A similar voting method is combined with deep learning and widely used in the field of object detection~\cite{votenet, xie2020mlcvnet, back_tracing} and instance segmentation~\cite{jiang2020pointgroup} from point clouds.
VoteNet~\cite{votenet} was the first method to combine traditional hough voting with deep learning methods, which utilized an MLP module to simulate voting procedure and predict offset coordinates and features for each seed point. The vote features and coordinates are grouped and aggregated to generate box proposals.
The voting method can also be used in 6-DOF pose estimation. PVN~\cite{peng2019pvnet} predicted the unit vector field for a given image and PVN3D~\cite{he2020pvn3d} predicted the keypoints offsets and instance semantic segmentation label for each pixel on a given RGBD image.
A similar strategy was also used for hand pose estimation from point clouds, Ge \etal~\cite{ge2018point} used a point-to-point regression method to predict the heatmap and unit vector field for each input hand point, which utilizes a similar strategy as voting.
To our best knowledge, we are the first attempt to use a voting module in the field of human mesh recovery, and we tailor this module to solve complex occlusions and noises when dealing with human mesh recovery in a single frame of partial point clouds.

\section{Methodology}

\subsection{Preliminaries}
\label{sub:preliminaries}

\subsubsection{3D Paramteric Human Model} 
\label{ssub:parametric_human_model}

The Skinned Multi-Person Linear Model (SMPL)~\cite{loper2015smpl} is used in this paper to represent 3D human bodies.
It includes two sets of parameters, \ie, the shape parameters $\vbeta\in\mathbb{R}^{10}$ for controlling height, weight, limb proportions, and the pose parameters $\vtheta\in\mathbb{R}^{3K}$ denotes the axis-angle representations of the relative 3D rotations of $K$ skeleton joints with respect to their parents in the kinematic tree.
To be specific, $K=23 + 1$, including one additional root joint indicates the global rotation of the human body model.
Given the pose and shape parameters, the SMPL model deforms a triangulated mesh template and generates a new mesh with $M=6890$ vertices, \ie, $\cM(\vtheta, \vbeta)$.
The deformation process $\cM(\vtheta, \vbeta)$ is differentiable \wrt $\vtheta$ and $\vbeta$, benefiting the integration of the SMPL model into a neural network with end-to-end training.
The SMPL model also provides $K$ body part labels associated with the skeleton joints, thus each vertex has a deterministic correspondence with its belonging body part.

\subsubsection{Deep Hough Voting for 3D Object Detection} 
\label{ssub:hough_voting_for_object_detection}

VoteNet~\cite{votenet} has shown great success in designing end-to-end 3D object detection networks for raw point clouds.
It reformulates the traditional Hough voting into a point-wise deep regression problem through shared MLPs.
VoteNet generates an object proposal by sampling the seed points from the input point cloud whose votes are within the same cluster.
It is more compatible with sparse point sets than region proposals and encourages proposal generation within adaptive receptive areas.
As it accumulates partial information to form confident detection, a voting-based approach would benefit skeleton discovery even with severe noises, missed areas, and occlusions.

\subsection{Overview}
\label{sub:overview}

As illustrated in Figure~\ref{fig:framework}, the input of our \method~is a point cloud $\mP\in\mathbb{R}^{N\times3}$, with a 3D coordinate for each of the $N$ points.
Such an input typically comes from depth sensors that scan partial views of the 3D human surfaces.
The outputs are two sets of SMPL parameters, one set includes the global parameters $\vphi_g = \{ \vbeta, \vtheta_0 \}$, and the other set has the local parameters $\vphi_l = \{ \vtheta_1, \ldots, \vtheta_{K-1} \}$.
The former controls the body shape and the global rotation, and the latter  indicates the relative rotations of the articulated skeleton joints.
These parameters deform a 3D human shape model $\cM(\vphi_g, \vphi_l)$ (mimicking the SMPL deformation process $\cM(\Theta, \vbeta)$) as a personalized triangulated mesh.
\method~consists of four main modules: (1) a partial vote generation and clustering module, (2) an occlusion-aware joint completion module, (3) a local parameter regressor, and (4) a global parameter regressor.
We will elaborate on these four modules in the following parts.

\subsection{Partial Vote Generation and Clustering}
\label{sub:partial_vote_generation_and_clustering}

From an input point cloud $\mP \in \mathbb{R}^{N\times3}$, we aim at generating $N$ votes, where each vote has a body part segmentation score vector $\mathbf{s}_i\in\mathbb{R}^{K}$, a 3D coordinate offset $\vo_i\in\mathbb{R}^3$ towards its nearest skeleton joint, and a high dimensional feature vector $\vf_i\in\mathbb{R}^{F_1}$.
These votes within the same segmentation results are then grouped into joint-level vote clusters, representing the skeleton joint of the corresponding segment.
There are three major steps: (1) point cloud feature learning through a backbone network, (2) simultaneously predicting three components of each vote, and (3) clustering votes for skeleton joint-level feature association.
Note that the input point clouds may only contain the partial human body scans, thus the output joint features may also partially fill the kinematic tree.

\subsubsection{Point Cloud Feature Learning}
\label{ssub:point_cloud_feature_learning}

%
We leverage PointNet++~\cite{qi2017pointnet++} as our feature extraction backbone due to its widespread success on point cloud analysis. 
Following 2D pose estimation~\cite{newell2016stacked}, this backbone network has multiple set abstraction (downsampling) layers and feature propagation (upsampling) layers with skip connections, thus encourage more contextual clues to enrich the geometric features.
It outputs the 3D point coordinate $\vx_i$ and a $F_1$-dimensional feature vector $\vb_i$ for each input point $\vp_i, i=1, \ldots, N$, while each point generates one vote.

\subsubsection{Partial Vote Generation}
\label{ssub:partial_vote_generation}

We apply a shared voting module network to generate the vote $\{ \mathbf{s}_i, \vo_i, \vf_i \}$ from each input base feature $\vb_i, i = 1, \ldots, N$ independently.
To be specific, the voting module consists of a multi-layer perceptron (MLP) network, which is followed by three independent heads for body part segmentation, offset regression, and feature updating, respectively.

The \textbf{segmentation head} employs a fully connected (FC) layer with \texttt{softmax} after the voting module to predict the body part segmentation score vector $\mathbf{s}_i\in\mathbb{R}^{K}$ for each input point $\vp_i$.
Our method employs the cross-entropy loss $\cL_\text{seg} = -\sum_{i=1}^N \log \mathbf{s}_i^{t_i}$ independently for each point to train the segmentation head.
$t_i$ is the ground truth (GT) body part label for the point $\vp_i$, $\mathbf{s}_i^{t_i}$ means the $t_i$-th entry of the score vector $\mathbf{s}_i$.
The GT segmentation label of each point is annotated by copying the body part label from the nearest vertex from the ground truth human shape model.

The \textbf{offset regression head} also employ an FC layer to output the Euclidean offsets $\vo_i \in \mathbb{R}^3$ for each input point $\vp_i$.
It aims to generate a vote offset towards the skeleton joint that the input point belongs to, so as to gather more valid contexts to reliably define the joint positions and their geometric features.
The predicted 3D offset $\vo_i$ is explicitly supervised by a regression loss
\begin{equation}
\cL_\text{vote-reg} = \frac{1}{N} \sum_{i}^N \sum_{k=0}^K \| \vo_i - (\vc^*_{k} - \vp_i) \|_\rho \cdot \mathbb{I}[\vp_i~\text{on body part}~k],
\end{equation}
where $\mathbb{I}[\vp_i~\text{on body part}~k]$ indicates whether an input point $\vp_i$ is on a body part $k$.
$\vc^*_k \in \mathbb{R}^3$ is the GT 3D coordinates for the $k$-th skeleton joint.
$\rho(\cdot)$ is the smooth-$\ell_1$ norm for robust regression.

The \textbf{feature updating head} uses a residual connection to update the vote features as $\vf_i = \vb_i + \Delta \vf_i, i = 1, \ldots, N$. $\Delta \vf_i$ is extracted by the shared MLP network with another independent FC layer.
The feature updating head does not receive explicit supervision.

The partial vote generation has an overall vote generation loss 
\begin{equation}
\label{eq:loss_vote_gen}
\cL_\text{vote-gen} = \lambda_\text{11} \cL_\text{vote-reg} +\lambda_\text{12} \cL_\text{seg},
\end{equation}
%
%
which is accompanied with subsequent losses for human mesh recovery.

\subsubsection{Vote Clustering}
\label{ssub:vote_clustering}

Given the predicted votes for all input points, we are ready to group and aggregate them into predicted skeleton joints and their associated features.
To be specific, we have the predicted position and feature for each joint as 
\begin{equation}
\vc_k = \frac{1}{\sum_{i=1}^N {\mathbf{s}_i^k}}\sum_{i=1}^N (\vp_i + \vo_i) \mathbf{s}_i^k, ~\text{and}~\vq_k = \frac{1}{\sum_{i=1}^N {\mathbf{s}_i^k}}\sum_{i=1}^N \vf_i \mathbf{s}_i^k.
\end{equation}
To be specific, we assume joints whose confidence score is below a pre-defined threshold as occluded and thus set the corresponding joint positions and features to be zero.

\subsection{Occlusion-aware Joint Completion}
\label{sub:occlusion_aware_joint_completion}

The input point clouds captured by depth sensors contain self-occlusions and missing areas, thus it is necessary to complete the skeleton joints in the kinematic tree for reliable pose estimation and global shape recovery.
In this part, we use an occlusion-aware joint completion module to fill in the missing joint features as well as the joint positions from the previous stage.

At first, we reorganize the joint features and positions of all skeleton joints as a tensor $\tilde{\mJ} \in \mathbb{R}^{K\times(3 + F_1)}$.
We first concatenate the joint position and feature in the channel dimensions and then list them in the order of the kinematic tree. 
Note that we set joints with a confidence score below a pre-defined threshold as occluded and thus set the corresponding joint positions and features to be zero, as indicated in Section~\ref{ssub:vote_clustering}. 
The occlusion-aware joint completion module infers the position and features for the occluded joints from the visible ones.
We use a two-layer MLP network to transform $\tilde{\mJ}$ into the completed joint feature $\tilde{\mQ} \in \mathbb{R}^{K\times F_2}$, and then an additional FC layer for predicting refined joint positions $\tilde{\mC} \in \mathbb{R}^{K\times3}$.
The completed 3D coordinates and features for all joints are fed into the global and local parameter regressors for SMPL-based HMR.

\subsection{Global Parameter Regression}
\label{sub:global_parameter_regression}

\begin{figure}[t]
\centering
\includegraphics[width=\linewidth]{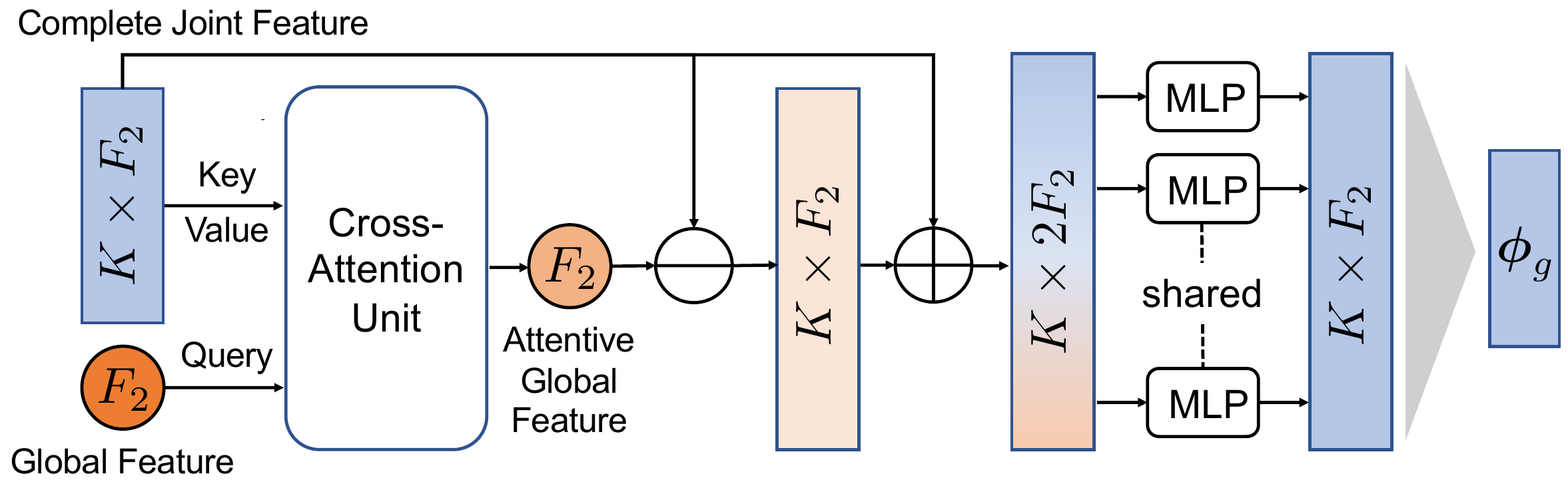}
\vspace{-7mm}
\caption{Global parameter regression module.}
\label{fig:global_parameter_regression}
\vspace{-3mm}
\end{figure}

The global parameter regression is to predict the shape parameter $\vbeta$ and the global rotation $\vtheta_0$.
%
%
We argue that the combination of global features and skeleton-based features can well describe the global shape information, \ie, the skeleton joints may partially describe the height, limb proportions and \emph{etc.}, while the holistic perception of the input point clouds can indicate the weight or size of the captured target.
To this end, a feature aggregation method similar to the edge convolution~\cite{wang2019dynamic} is adopted to obtain a skeleton-aware global parameter regression, as shown in Figure~\ref{fig:global_parameter_regression}.
The inputs of this module are the completed joint features $\tilde{\mQ} \in \mathbb{R}^{K\times F_2}$, and the global feature $\vg \in \mathbb{R}^{1\times F_2}$ extracted at the bottleneck layer of the backbone network for point cloud feature learning (shown in Section~\ref{ssub:point_cloud_feature_learning}).
Since the global features and the local joint features are heterogeneous, we apply a cross-attention module~\cite{vaswani2017attention} to align them, where the query is the global feature $\vg$, the key and value are the joint features $\tilde{\mQ}$, thus the attentive global feature is 
\begin{equation}
\tilde{\vg} = \mathtt{softmax}\left( {\vpsi_Q(\vg) \vpsi_K(\tilde{\mQ})^\top}/{\sqrt{F_2}} \right) \vpsi_V(\tilde{\mQ}).
\end{equation}
$\vpsi_Q(\cdot), \vpsi_K(\cdot), \vpsi_V(\cdot)$ are three projection layers for query, key and value, respectively.
Then the edge convolution operation is performed between $\tilde{\vg}$ and each joint feature $\tilde{\vq}_i$, $i=0,\ldots,K$, written as $\vpsi_\text{EdgeConv}([\tilde{\vq}_i; \tilde{\vg} - \tilde{\vq}_i])$, where $\vpsi_\text{EdgeConv}(\cdot)$ is one MLP layer that is shared for each pair.
The resultant features are further fed into another FC layer to predict the global parameters $\vphi_g \in \mathbb{R}^{19}$, including $\vtheta\in\mathbb{R}^{10}$ and a vectorized root rotation matrix $\text{vec}(\mR(\vtheta_0))$, following the recent successes~\cite{pavlakos2018learning, omran2018neural, zhang2020learning}.
%
%

\subsection{Local Parameter Regression}
\label{sub:local_parameter_regression}

The local parameter regression is to predict the relative rotations $\{\vtheta_k \}_{k=1}^{K-1}$ of the articulated skeleton joints.
We employ two layers graph convolution network (GCNs) onto the skeleton graph to incorporate contextual relations, where the graph nodes are the completed joint features $\tilde{\mQ}$, the graph edges are the edges in the kinematic tree.
The output is the vectorized rotation matrices for all skeleton joints except that of the root joint.
%
Thus the local parameters take the format of $\vphi_l \in \mathbb{R}^{(K-1)\times9}$.

\subsection{Losses}
\label{sub:losses}

The overall loss function for our \method~can be divided into the vote generation loss $\cL_\text{vote-gen}$ as defined in Equ.~\eqref{eq:loss_vote_gen}, the parameter regression loss for the global and local parameters, and the model fitting loss, written as
\begin{equation}
\cL = \lambda_\text{1} \cL_\text{vote-gen} + \lambda_\text{2} \cL_\text{param-reg} + \lambda_\text{3} \cL_\text{model-fit}.
\end{equation}
$\cL_\text{param-reg}$ and $\cL_\text{model-fit}$ will be depicted in the following parts.

\subsubsection{Parameter Regression Loss}
\label{ssub:parameter_regression_loss}
In order to achieve higher pose prediction accuracies, following~\cite{pavlakos2018learning, omran2018neural, zhang2020learning}, we directly predict the rotation matrix $\mR_k \in \mathbb{R}^{3\times3}, k=0,\ldots,K$ for each joint, instead of its axis-angle representation.
Therefore, We apply SMPL parameter regression loss to SURREAL dataset with ground truth SMPL parameters, which is the combinational losses for shape and pose, $\cL_\text{SMPL} = \| \vbeta - \vbeta^* \|_1 + \sum_{k=0}^{K-1} \| \mR_k - \mR_k^* \|_1$.
To ensure the orthogonality of the generated rotation matrices, we explicitly include an orthogonality regularization for the predicted rotation matrices, such as $\cL_\text{orth} = \sum_{k=0}^{K-1} \| \mR_k \mR_k^\top - \mI \|_2$.
Therefore, the parameter regression loss is $\cL_\text{param-reg} = \lambda_\text{21}\cL_\text{SMPL} + \lambda_\text{22}\cL_\text{orth}$.

\subsubsection{Model Fitting Loss}
\label{ssub:model_fitting_loss}

The model fitting loss includes three parts. The first one is the skeleton fitting loss, \ie, $\cL_\text{skeleton} = \frac{1}{K}\sum_{i=0}^K \| \vc_k - \vc_k^* \|_1$, which matches the predicted joint positions and the GT joint positions without occlusions.
The second loss is the vertex fitting loss, which is implemented as the per-vertex Euclidean error between the predicted human shape model and the GT model, such as $\cL_\text{vertex} = \frac{1}{M}\| \cM(\vphi_g, \vphi_l) - \mM^* \|_1$, where $\mM^*$ indicate the 3D coordinates of the vertices of the GT human model, $M$ is the number of vertices in the SMPL model.
The third loss is the half-term chamfer distance between the predicted human shape model and the input point cloud, \ie, $\cL_\text{CD} = \frac{1}{N}\sum_{i=1}^N \min_{j = 1, \ldots, M} \| \vp_i - \vm_j \|_2$.
It finds the closest vertex in the predicted model for each point $\vp_i$ in the input point cloud.
$\vm_i\in\mathbb{R}^3$ is the 3D coordinate of the $i^\text{th}$ vertex.
Therefore, the overall model fitting loss is $
\cL_\text{model-fit} = \lambda_\text{31} \cL_\text{vertex} + \lambda_\text{32}\cL_\text{CD} + \lambda_\text{33} \cL_\text{skeleton}$.

\section{Experiments}


\subsection{Setup}
\label{ssub:datasets_and_evaluation_metrics}

\subsubsection{Datasets}
\label{ssub:datasets}

We conduct our experiments on two synthetic datasets, \ie, SURREAL~\cite{varol2017learning} and DFAUST~\cite{bogo2017dynamic}, and a real-world dataset, Berkeley MHAD~\cite{ofli2013berkeley}.
\method~ is trained on SURREAL and DFAUST datasets since they provide GT SMPL models for each input point cloud.
The Berkeley MHAD dataset is applied for testing the reconstruction performance in the real scenario. 

\noindent
\textbf{- SURREAL~\cite{varol2017learning}}.
The training set of the SURREAL dataset contains $55,001$ depth clips of synthetic human motions. 
We follow the strategy of Wang~\etal~\cite{wang2020sequential} to uniformly sample $1,000,000$ depth images as the training set, and $10,000$ frames as the testing set.
We re-projected the depth images to obtain the partial point clouds.

\noindent
\textbf{- DFAUST~\cite{bogo2017dynamic}.}
It contains $40,000$ real undressed human meshes for $65$ action sequences. The training set of DFAUST is constructed by projecting human meshes from the former $55$ action sequences onto $10$ camera views, and then randomly sampling $250,000$ depth images.
The testing set is sampled from the latter $10$ action sequences, and also randomly samples $10,000$ frames as the testing set, following the setting in~\cite{wang2020sequential}.
We apply the same point cloud generation strategy as that for the SURREAL dataset.

\noindent
\textbf{- Berkeley MHAD~\cite{ofli2013berkeley}}.
This dataset is composed of two-view depth sequences captured from two synchronized Kinect cameras. It has a wide range of poses, containing 12 action sequences performed by 12 actors.
Berkeley MHAD also provides per-frame 3D human keypoints annotations, which are also used to generate bounding boxes to crop out the human regions. We sample about $2400$ frames from the first eight sequences to serve as the test set. Note that two depth views of the same action frame are both converted into point clouds as two separate test samples.

\subsubsection{Evaluation Metrics}
\label{ssub:evaluation_metrics}

We consider several evaluation metrics to validate the performance of our method in different aspects.
For synthetic datasets that contain GT human models, we use the widely adopted per-vertex error (PVE) to compute the average reconstruction error between the corresponding vertices from two SMPL models.
We also report the maximum PVE (PVE max) error to evaluate the robustness of our method when dealing with extreme poses.
Moreover, we also calculate the mean per-joint position error (MPJPE) to indicate the prediction accuracy of the 3D joint locations.
For real datasets that we do not have GT human models, we report the average Chamfer Distance (CD) as described in Section~\ref{sub:losses}.





\subsubsection{Implementation Details}
\label{ssub:implementation_details}

We randomly sample the raw point clouds to a fixed size with $N = 2,500$ as input. 
%
The PointNet++ backbone consists of $4$ layers of set abstraction and feature propagation modules, respectively.
The dimension of the per-point feature from the last layer of PointNet++ is $128$, the dimension of the cluster joint feature is $131$.
In the partial vote generation and clustering stage, we empirically set the occluded pre-defined threshold as $0.1$.
In the training process, we use the Adam optimizer~\cite{kingma2015adam} with a learning rate of $1\mathrm{e}{-4}$. 
We set the loss weights of first three loss terms as $\lambda_\text{1} = 1 $ and $\lambda{2} = \lambda{3} = 10$. 
For vote generation loss, loss weights are set as $\lambda_\text{11} = 0.1$ and $\lambda_\text{12} = 1$. We set $\lambda_{21} = 1$ when training SURREAL, and $\lambda_{21} = 0$ for training DFAUST. 
The rest loss weights are all set to be $1$. 
During testing, we downsample the predicted shape model to $1723$ vertices for the calculation of PVE and PVE max, so as to fairly compare with the baseline method~\cite{wang2020sequential}.


\begin{table}[t]
\caption{Per vertex reconstruction errors (mm) of different methods tested on synthetic datasets.}
\vspace{-3mm}
\label{tab:comparison_synthetic_datasets}
\begin{tabular}{c|c|c}
\hline
\hline
Method & SURREAL~\cite{varol2017learning} & DFAUST~\cite{bogo2017dynamic} \\
\hline
HMR~\cite{kanazawa2018end} & 75.85 & 76.87 \\
GraphCMR~\cite{kolotouros2019convolutional} & 73.81 & 75.31 \\
3DCODED~\cite{groueix20183d} & 41.8 & 43.5 \\ 
Skeleton~\cite{jiang2019skeleton} & 80.5 & 82.6 \\
Sequential (Non Param.)~\cite{wang2020sequential} & 21.2 & 22.2 \\
Sequential (Param.)~\cite{wang2020sequential} & 24.3 & 25.2\\
\hline
Our Method & \textbf{20.2} & \textbf{21.5} \\
\hline
\hline
\end{tabular}
\end{table}

\begin{figure}[t]
\centering
\includegraphics[width=\linewidth]{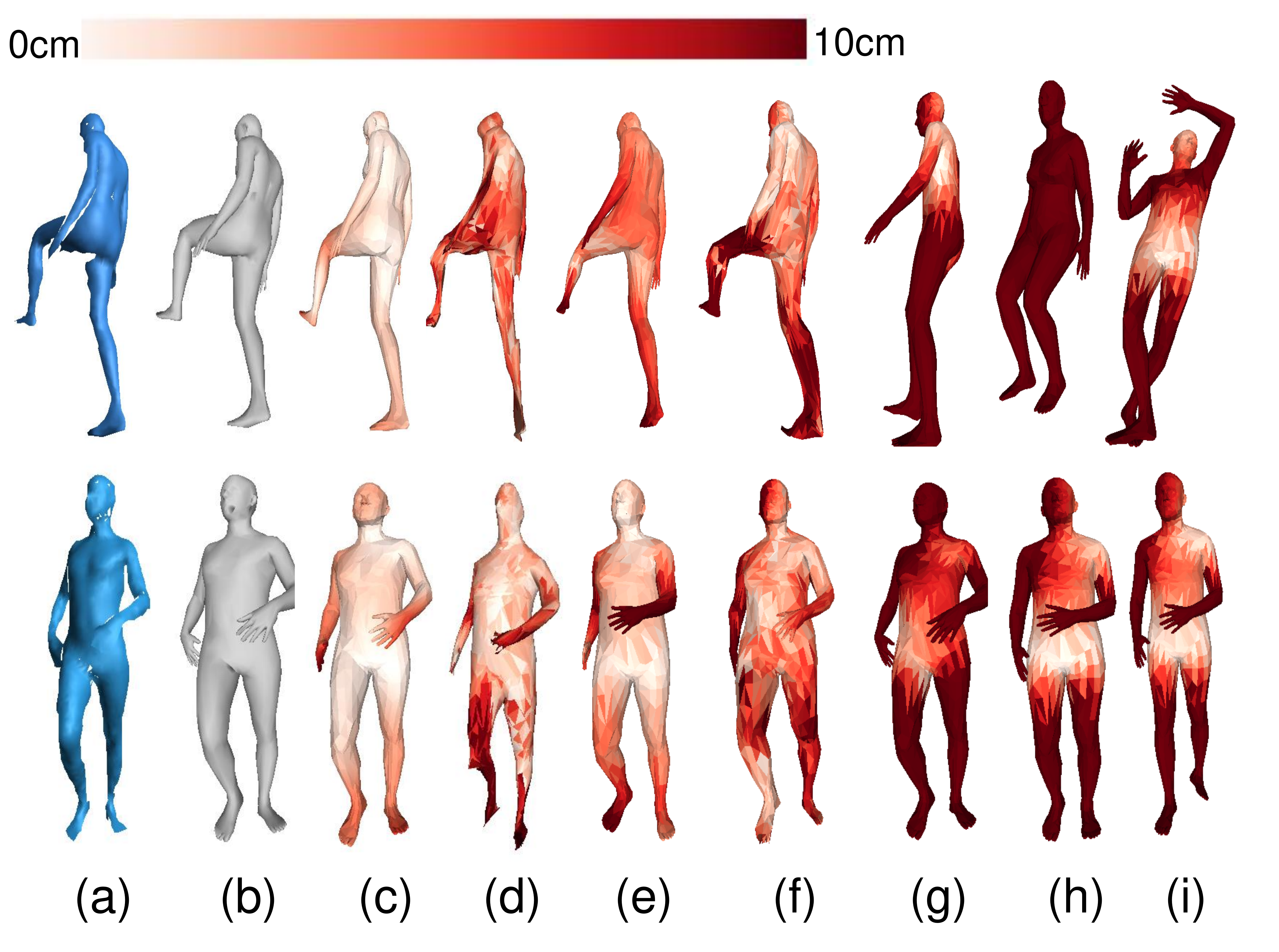}
\vspace{-0.5cm}
\caption{Qualitative comparison of different methods on the SURREAL dataset~\cite{varol2017learning}. (a) The input partial point clouds, visualized as partially visible meshes. (b) The ground truth human meshes. From (c) to (i), the reconstructed results by (c) \method, (d) the non-parametric and (e) parametric variants of Wang~\etal~\cite{wang2020sequential}, (f) 3D-CODED~\cite{groueix20183d}, (g) Jiang~\etal~\cite{jiang2019skeleton}, (h) GraphCMR~\cite{kolotouros2019convolutional} and (i) HMR~\cite{kanazawa2018end}. The color indicates the degree of reconstruction error in each vertex.}
\label{fig:comparison_surreal}
\end{figure}

\subsection{Comparison to State-of-the-art Methods}
\label{sub:comparison_to_state_of_the_art_methods}

\subsubsection{Comparison on Synthetic Data}
\label{ssub:comparison_on_synthetic_data}

In this part, we report the quantitative and qualitative comparisons between our method and the reference methods in Table~\ref{tab:comparison_synthetic_datasets} and Figure~\ref{fig:comparison_surreal}. The quantitative evaluations are both conducted on the SURREAL dataset~\cite{varol2017learning} and the DFAUST dataset~\cite{bogo2017dynamic}. The qualitative evaluations are performed on the SURREAL dataset.
The reference methods are all retrained on these datasets for a fair comparison.

We first compare \method~with two depth image-based human mesh recovery methods by retraining the recently RGB-based HMR~\cite{kanazawa2018end} and GraphCMR~\cite{kolotouros2019convolutional} methods onto depth data.
%
%
Both the quantitative and qualitative comparisons demonstrate that our method outperforms them by a large margin.
%
%
It shows that the point clouds have unique benefits towards 3D human shape reconstruction, in addition to data in the 2D modality.

We also compare our model with 3D human mesh recovery methods upon the point clouds~\cite{wang2020sequential, groueix20183d, jiang2019skeleton}.
%
%
To fairly compare our method with them, we add the SMPL parameter supervision to Jiang~\etal~\cite{jiang2019skeleton}, and reproduce a parametric version of Wang~\etal~\cite{wang2020sequential} through an optimization-based SMPL fitting on predicted vertices. Note that we just employ the single frame-based framework of Wang~\etal~\cite{wang2020sequential}.
The quantitative results demonstrate that the proposed~\method~tremendously reduces reconstruction errors than these reference methods.
The visualization comparisons show that Jiang~\etal~\cite{jiang2019skeleton} produce results with large pose mismatches, especially the global poses, the non-parametric version of Wang~\etal~\cite{wang2020sequential} distorts the human shapes when facing occlusion or hard poses.
3D-CODED~\cite{groueix20183d} can also roughly reconstruct the human shapes, but its results may suffer from rugged or over-bent meshes, as shown in Figure~\ref{fig:comparison_surreal}.

\begin{figure}[t]
\centering
\includegraphics[width=0.95\linewidth]{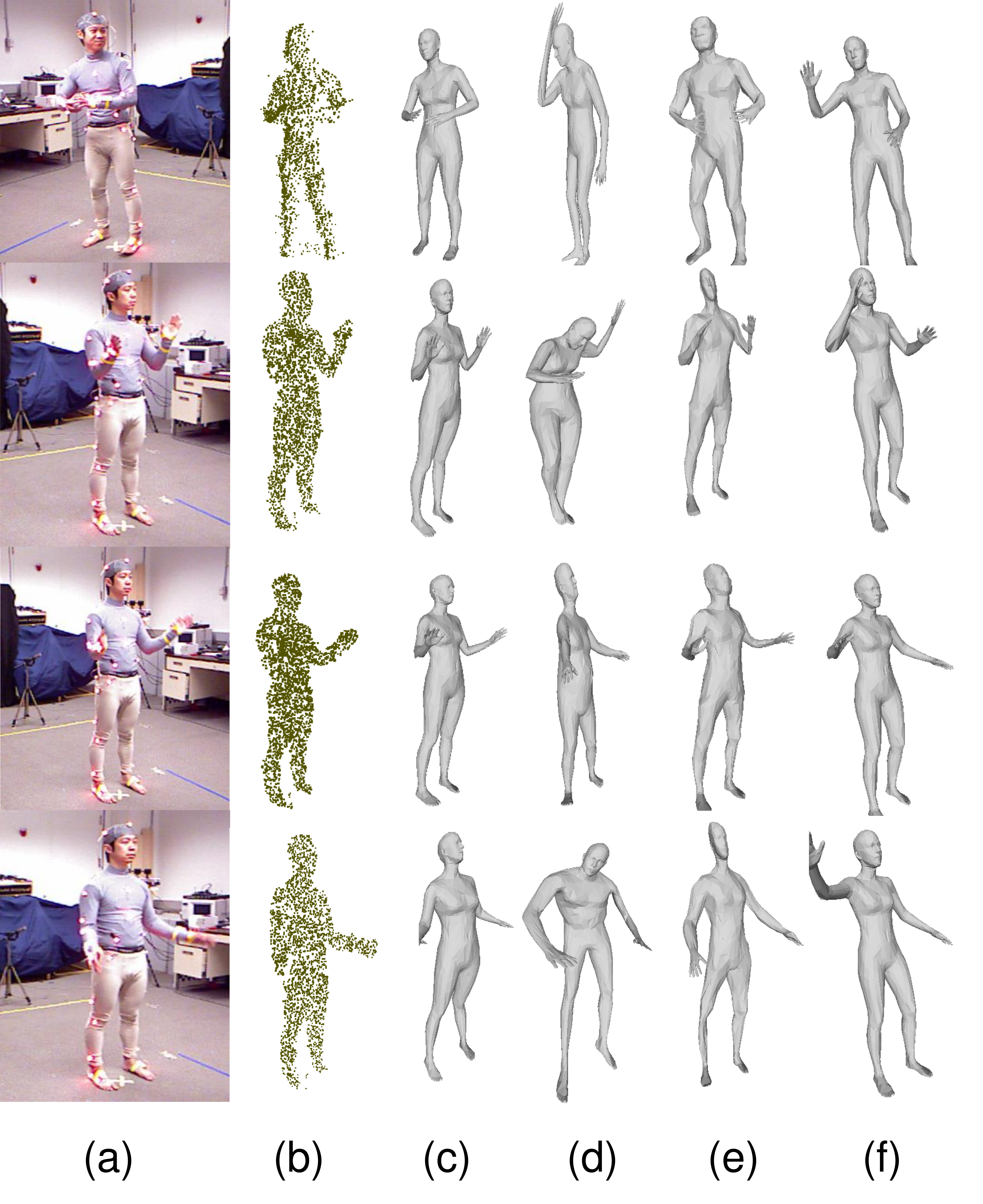}
\vspace{-0.5cm}
\caption{Reconstruction results on different samples from the Berkeley MHAD dataset~\cite{ofli2013berkeley}, without the weakly supervised fine-tuning scheme~\cite{wang2020sequential}. (a) and (b) shows the input point clouds and their corresponding RGB images. (c) are the predicted meshes of our~\method. (d) to (f) are the results by Wang~\etal~\cite{wang2020sequential}, 3D-CODED~\cite{groueix20183d}, and Jiang~\etal~\cite{jiang2019skeleton}.}
\label{fig:comparison_real_data}
\end{figure}

\subsubsection{Comparison on Real Data}
\label{ssub:comparison_on_real_data}

\begin{table}[t]
\caption{Mean Chamfer Distance (mm) of different methods tested on MHAD dataset. Initial and Final suggest results before and after weakly supervised fine tuning.}
\vspace{-0.2cm}
\label{tab:comparison_mhad_dataset}
\begin{tabular}{l|c|c}
\hline
\hline
Method & Initial & Final \\
\hline
3DCODED~\cite{groueix20183d} & 56.29 & 37.62 \\ 
Jiang~\etal~\cite{jiang2019skeleton} & 121.63 & 29.28 \\
Wang~\etal~\cite{wang2020sequential} & 81.29 & 36.79 \\
\hline
Our Method & \textbf{51.76} & \textbf{24.44} \\
\hline
\hline
\end{tabular}
\end{table}

We also evaluate the generalization ability of our~\method~on the real data, in comparison to the reference methods.
In this study, all methods are tested on the Berkeley MHAD dataset~\cite{ofli2013berkeley} but trained on the SURREAL dataset~\cite{varol2017learning} and DFAUST~\cite{bogo2017dynamic}.
As visualized in Figure~\ref{fig:comparison_real_data}, the proposed method achieves more reliable reconstructions than the reference methods, even though the input point clouds are heavily contaminated by occlusions and noises.
Note that we report both results before and after the weakly supervised fine-tuning scheme as frequently used in the previous methods~\cite{jiang2019skeleton,wang2020sequential,groueix20183d}, the reported results before the weakly supervised fine-tuning can better indicate the capabilities of different feed-forward networks of these methods, without the interference of the aforementioned post-processing.
 Without the weakly supervised fine-tuning scheme, the methods by Wang~\etal~\cite{wang2020sequential} and Jiang~\etal~\cite{jiang2019skeleton} usually predict incorrect poses, manifesting that their methods are sensitive to the domains of the training data, and much less robust to changes in the point clouds.
We also report the results after weakly supervised fine-tuning results used in previous methods in \cite{wang2020sequential}. The overall weakly supervised loss term including a Chamfer distance loss to fit the mesh to input point clouds, a Laplacian loss to preserve surface smoothness, an edge loss to penalize unnatural edges and enforce length consistency. We optimize the overall loss term on MHAD for about 100 epochs. Comparison results are also shown in Table~\ref{tab:comparison_mhad_dataset}.
In this case, the proposed method still achieves significantly better performances than other methods.

Our method may fail in cases of rare poses, while the reference methods may also fail as well.
We also notice that our method often fails to predict correct shape parameters for real data, one possible reason may be the limited shape variance of the synthetic training dataset, while the noise in the MHAD~\cite{ofli2013berkeley} data also adds additional difficulties to the shape estimation.


\begin{table}[t]
\centering
\caption{Ablation study of \method~on the SURREAL dataset~\cite{varol2017learning}. The performances are evaluated on the PVE, MAJPE and PVE max, respectively.}
\vspace{-3mm}
\label{ablation}
\begin{tabular}{l|c|c|c}
\hline
\hline
Method & PVE & MPJPE & PVE max  \\
\hline
Ours & 20.14 & 17.7 & 220.78 \\
\hline
w/o Voting & 57.51 & 49.2 & 678.4 \\
w/o Joint Completion & 59.44 & 49.47 & 444.7 \\
w/o Global Parameter Regressor & 26.2 & 22.93 & 275.9 \\
\hline
\hline
\end{tabular}
\label{table:ablation_study}
\end{table}

\subsection{Ablation Study}
\label{sub:ablation_study}

We conduct a list of ablation studies of our~\method, the PVE, MPJPE and PVE max results are reported in Table~\ref{table:ablation_study}. 

\subsubsection{Partial Vote Generation and Clustering}
\label{ssub:partial_vote_generation_and_clustering}

\begin{figure}[t]
\centering
\includegraphics[width=0.9\linewidth]{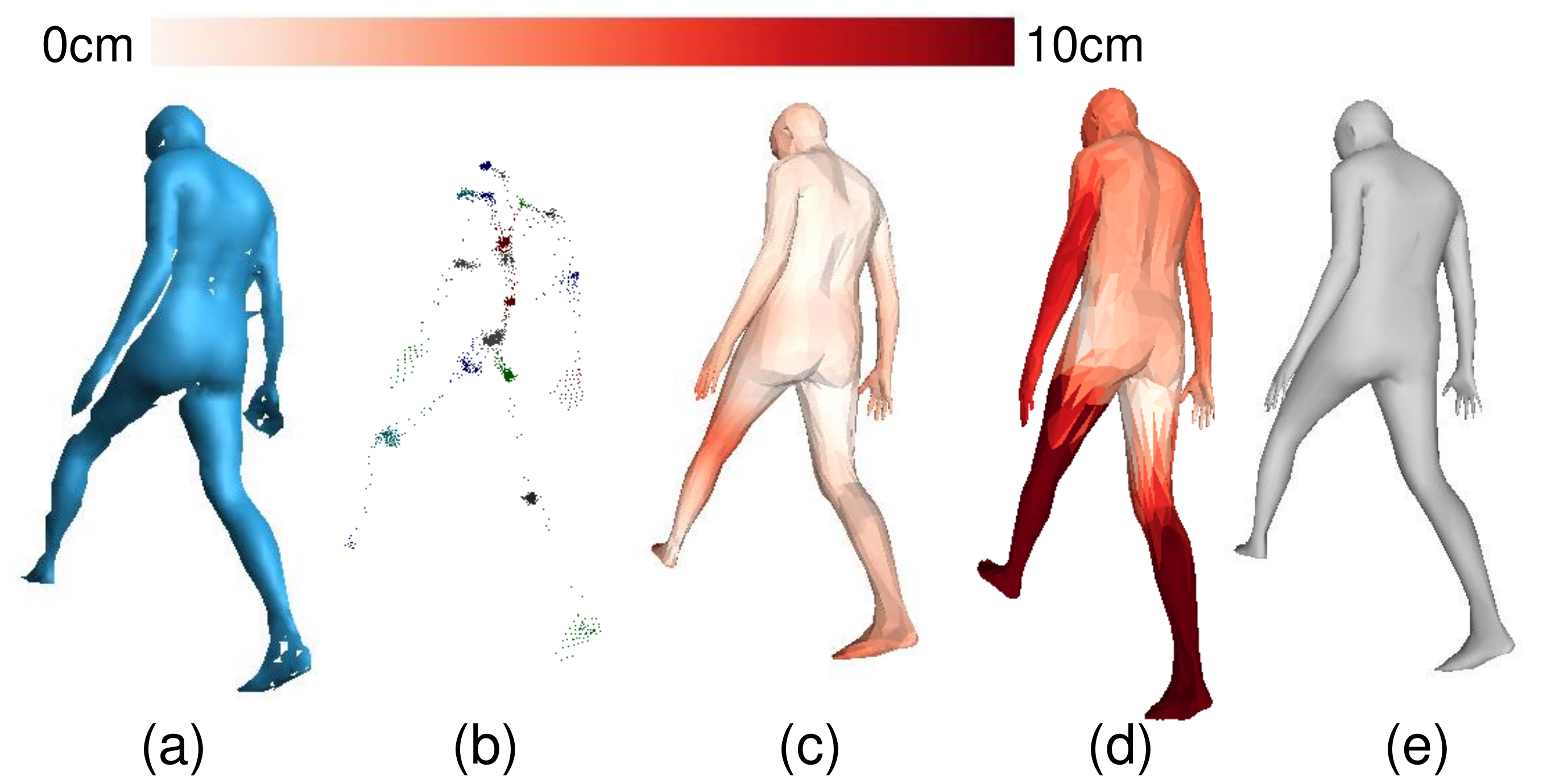}
\vspace{-0.2cm}
   \caption{Reconstruction accuracies with and without the Partial Vote Generation and Clustering module. (a) is the point cloud input. (b) is the vote results. (c) is the reconstruction error with the Partial Vote Generation and Clustering module. (d) is the reconstruction error without this module. (e) is the ground truth human mesh.}
\label{novoting}
\vspace{-0.2cm}
\end{figure}

We first evaluate the effectiveness of our partial vote generation and clustering module by creating a baseline method named ``w/o voting'', by replacing these two modules with the skeleton-aware attention module as in~\cite{jiang2019skeleton}. 
In ``w/o voting'' baseline, the pointnet++ downsamples point cloud to $64$ seed point features, and an attention module is used to map unordered seed point features into skeleton joint features.
The skeleton joint features are further fed to the joint completion module for feature refinement. 
%
%
The results in Table~\ref{ablation} and Figure~\ref{novoting} show that our method with partial vote generation and clustering can produce results with much lower errors than this baseline method does.
Compared to attention-based solutions, our method can capture a more accurate skeleton structure, thus benefiting from a more reliable pose estimation. Note that we also report the maximum PVE value during evaluation. \method~without voting module has a much larger maximum reconstruction error on the SURREAL dataset, indicating the robustness of our partial vote generation and clustering module on hard poses.

\begin{figure}[t]
\centering
\includegraphics[width=0.9\linewidth]{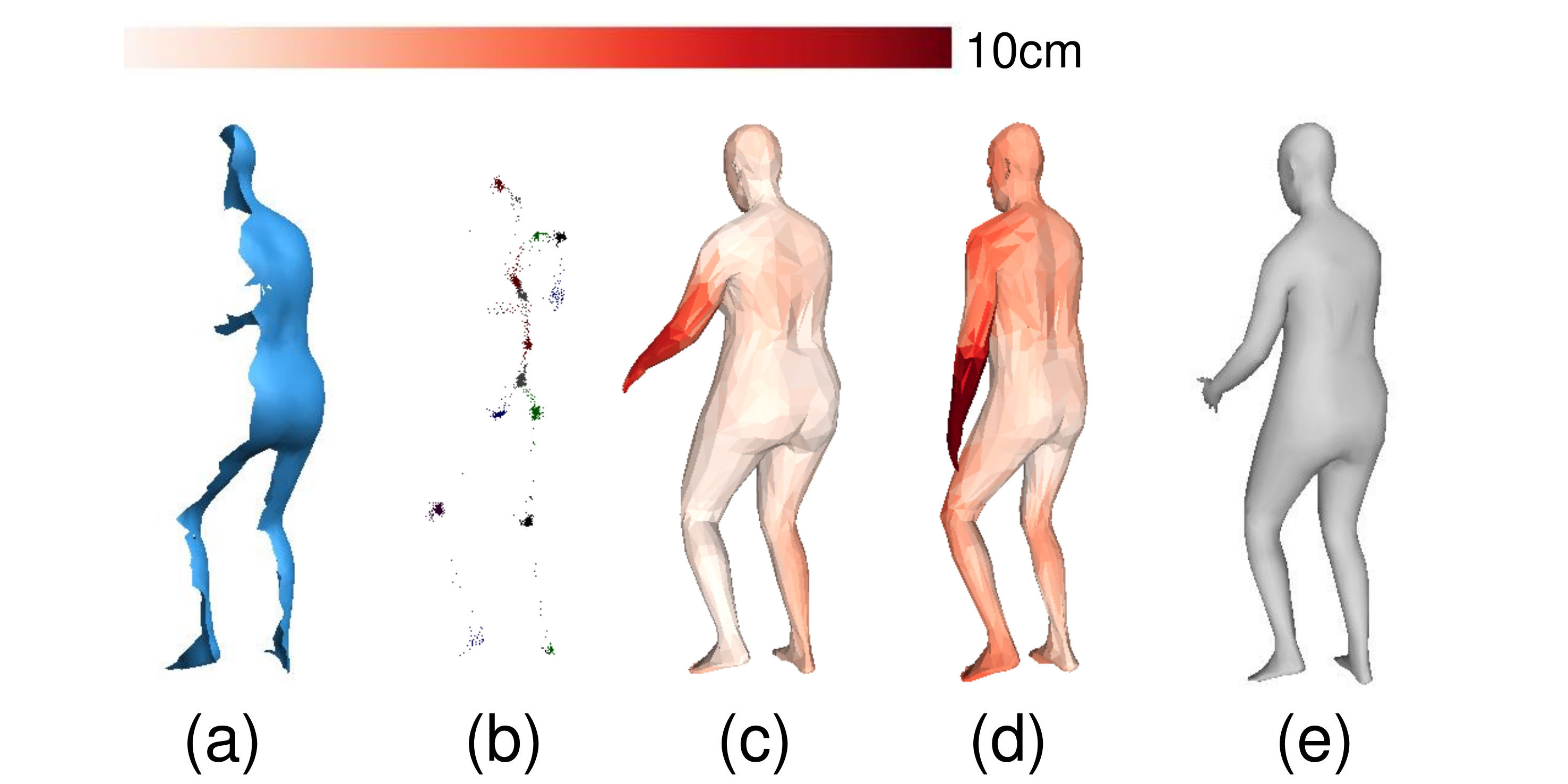}
\vspace{-0.3cm}
   \caption{Reconstruction accuracies with and without the Occlusion-aware Joint Completion module. (a) is the point cloud input. (b) is the vote results. (c) is the reconstruction error with the Occlusion-aware Joint Completion module. (d) is the reconstruction error without this module. (e) is the ground truth human mesh.}
\label{nocompletion}
\vspace{-0.2cm}
\end{figure}

\subsubsection{Occlusion-aware Joint Completion}
\label{ssub:occlusion_aware_joint_completion}

To evaluate the effectiveness of our occlusion-aware joint completion module, we also conduct a baseline ``w/o joint completion'' by feeding incomplete joint coordinates and features into the global and local parameter regressors.
The comparison results of all evaluation metrics in Table~\ref{ablation} demonstrate that our joint completion module is essential to achieve better performances.
As shown in Figure~\ref{nocompletion}, the occluded hand fails to be accurately recovered without the joint completion module. In contrast, our \method~can better handle the occlusions and guess the most likely human poses from the partial input point clouds.

\subsubsection{Global Parameter Regression}
\label{ssub:global_parameter_regression}

\begin{figure}[t]
\centering
\includegraphics[width=0.85\linewidth]{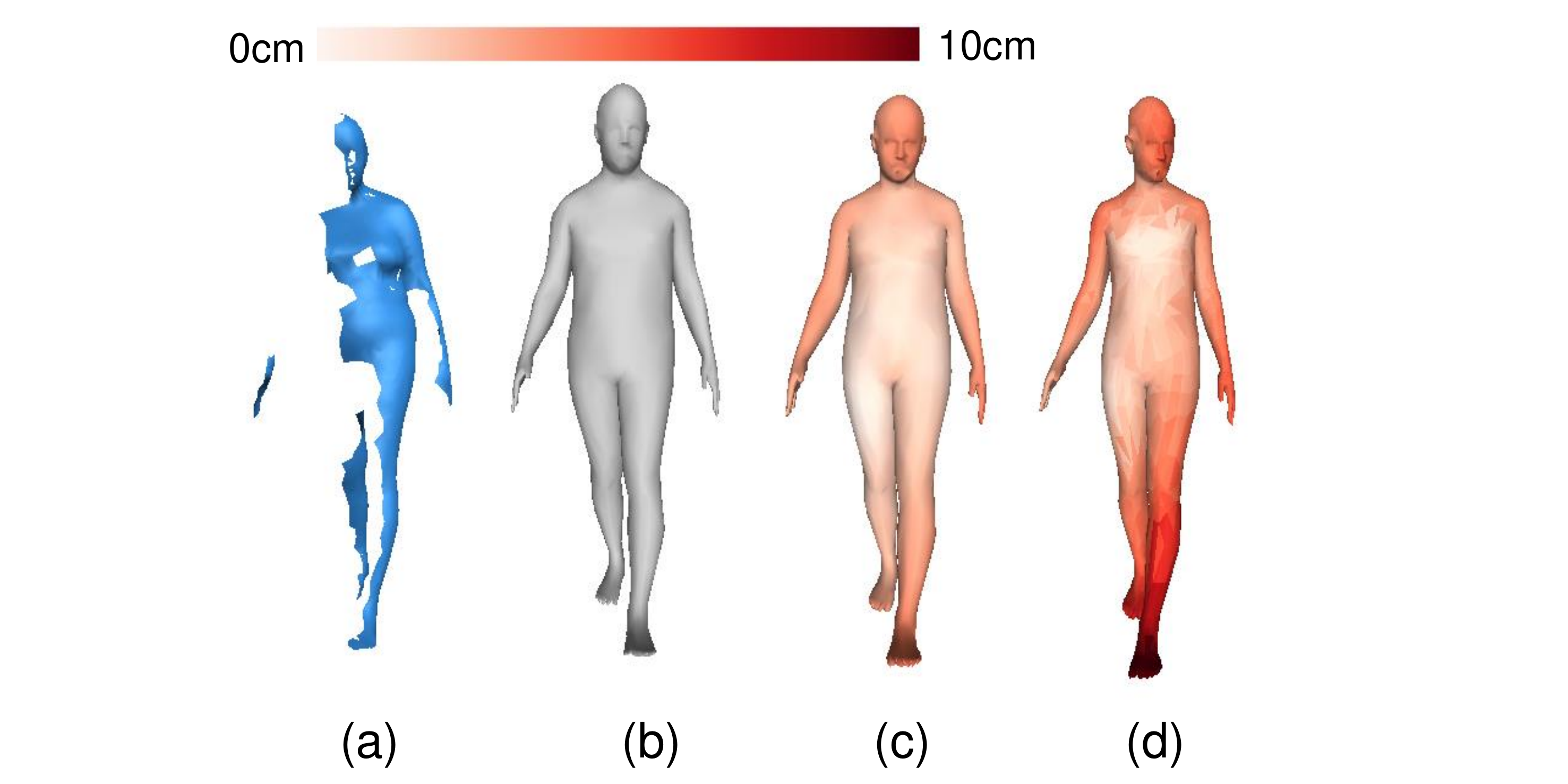}
\vspace{-0.4cm}
   \caption{Reconstruction accuracies with and without the Global Parameter Regression. (a) is the point cloud input. (b) is the GT mesh. (c) is the reconstruction error with the Global Parameter Regression module. (d) is the reconstruction error without this module.}
\label{noedgeconv}
\end{figure}

We further evaluate the effectiveness of our global parameter regression.
For baseline method ``w/o global parameter regression'', we apply a single layer MLP on top of the concatenated complete joint coordinates and features to estimate the SMPL global parameters~$\vphi_g \in \mathbb{R}^{19}$.
The reconstruction errors are listed in Table~\ref{ablation}. As shown in Figure~\ref{noedgeconv}, the result without global parameter regression estimates giant shape error compared to the result with our complete method. 

\begin{figure}[t]
\centering
\includegraphics[width=0.5\linewidth]{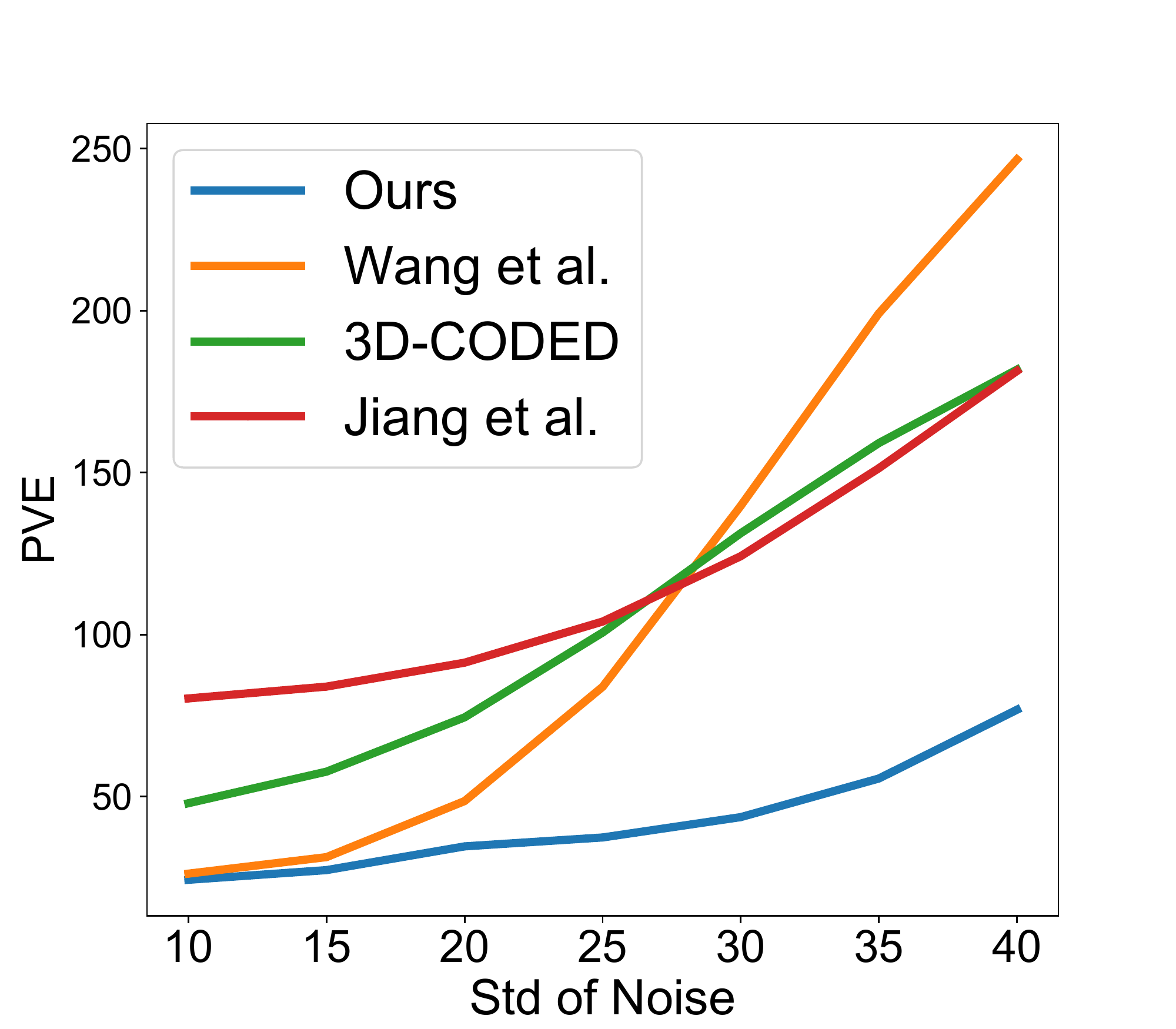}\includegraphics[width=0.5\linewidth]{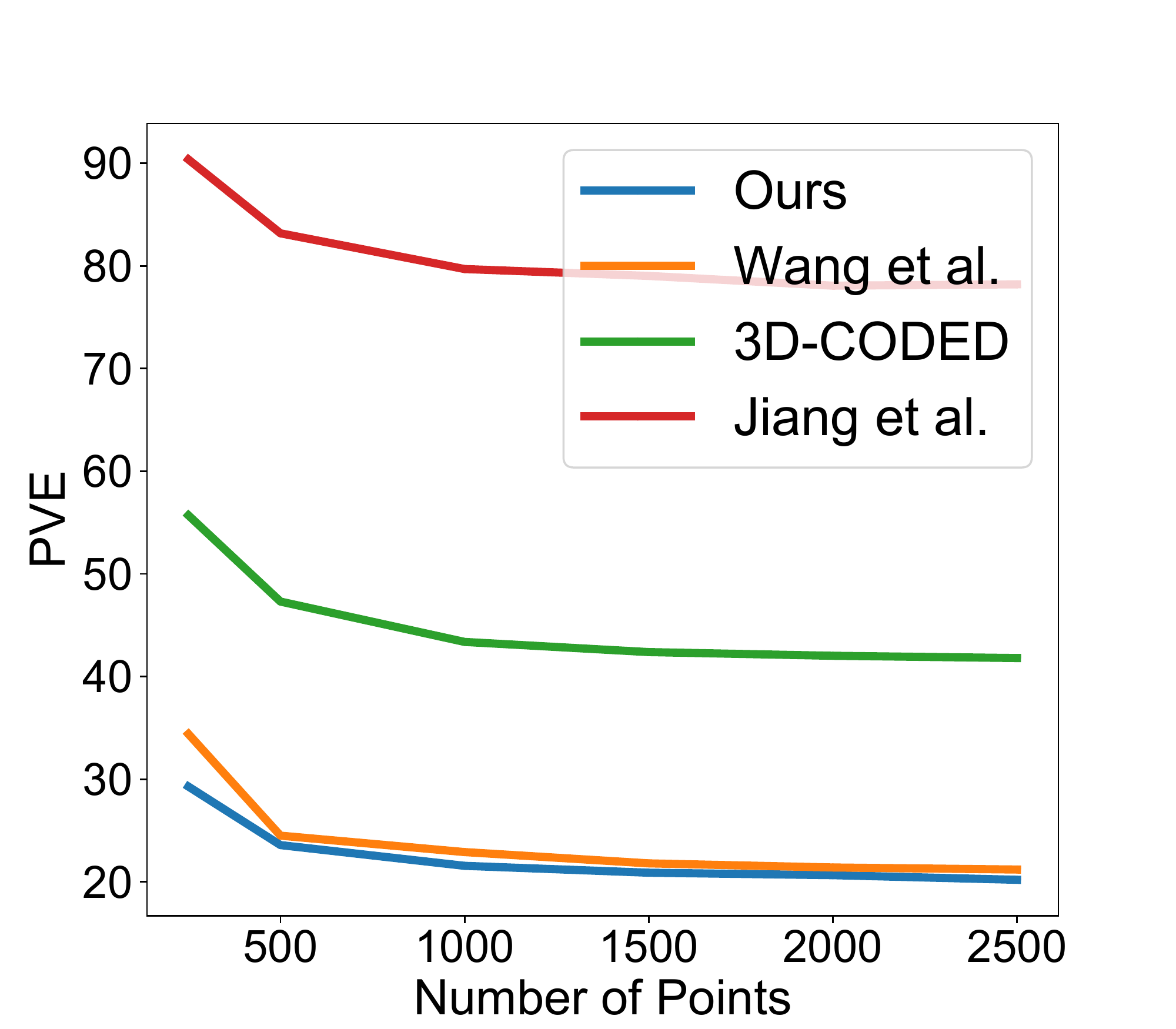}
\vspace{-0.4cm}
   \caption{Per Vertex Error of point cloud based methods against varying noise levels (left), and varying numbers of points (right) on the SURREAL testset.}
\label{noise_figure}
\end{figure}

\begin{figure}[t]
\centering
\includegraphics[width=\linewidth]{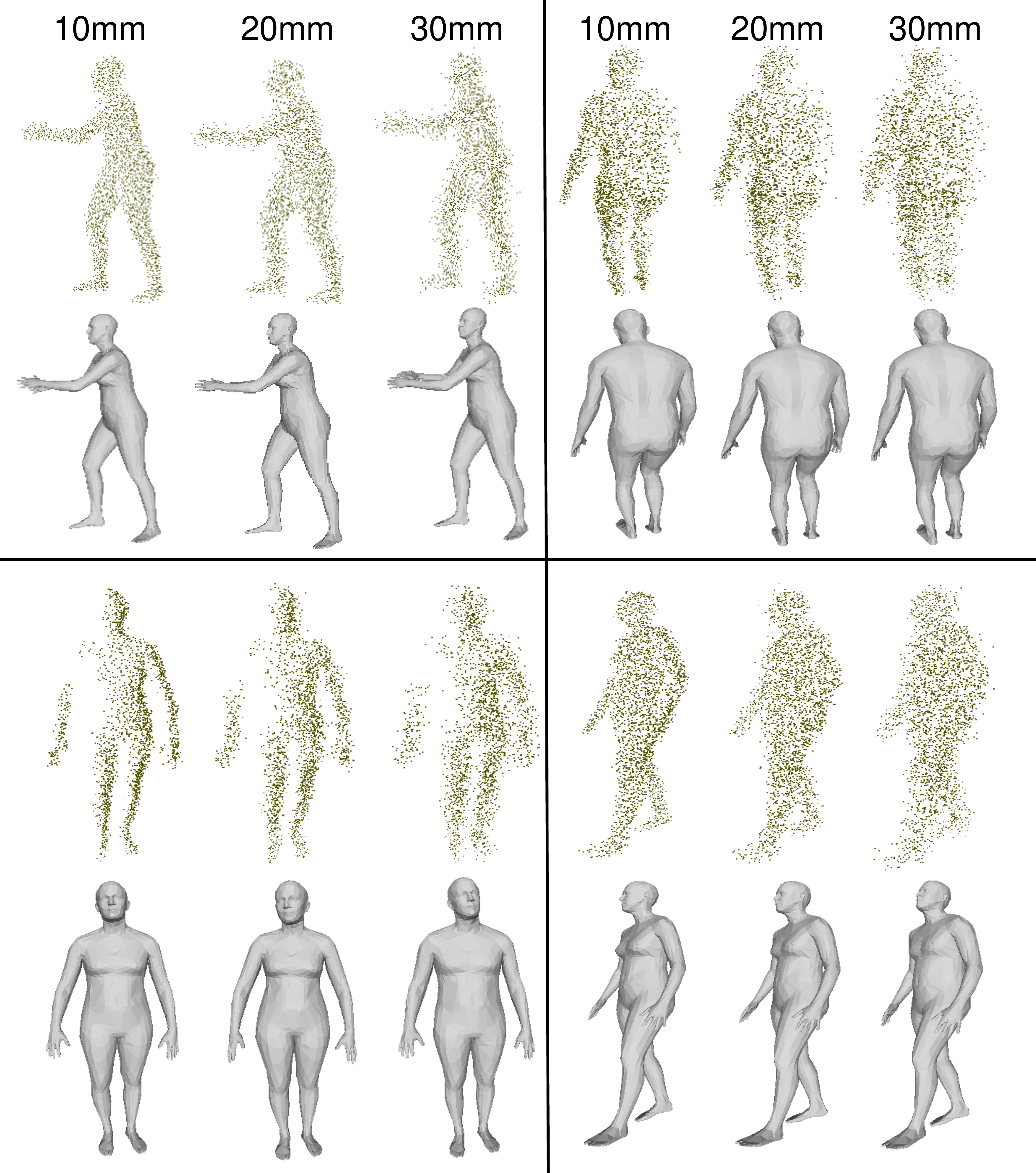}
\vspace{-3mm}
\caption{Qualitative results of our VoteHMR under different noise variation levels in the input point clouds. For each result, from left to right, we show the reconstruction results by applying our model to inputs with noise standard deviation of $10$mm, $20$mm, $30$mm, respectively.}
\label{noise_visualization}
\end{figure}

\subsection{More Experiments about Robustness}
\label{sub:robustness_exp}

\subsubsection{Reconstruction against Varying Noise Levels}

Figure~\ref{noise_figure} (left) compares different methods about the PVEs at different noise levels on the SURREAL test set.
In this experiment, each point in a test sample is added a random 3D noise offset according to a given standard deviation.
We retrain our \method~and the previous methods~\cite{wang2020sequential, groueix20183d, jiang2019skeleton} with a noise standard deviation of $10$mm. During the evaluation stage, we increase the noise standard deviation of the test set to up to $40$mm.
The performance of Wang~\etal~\cite{wang2020sequential} drops dramatically as the standard deviation of the added noises increase.
While the proposed methods~\method~ usually present the lowest PVE scores in all the noise levels.
We also visualize the results of our method in Figure~\ref{noise_visualization}, where different levels of noises do not significantly distort the pose predictions. The increase of PVE scores is mainly caused by inevitable prediction errors of shape parameters due to large noise variance.

%

\begin{figure}
\centering
\includegraphics[width=\linewidth]{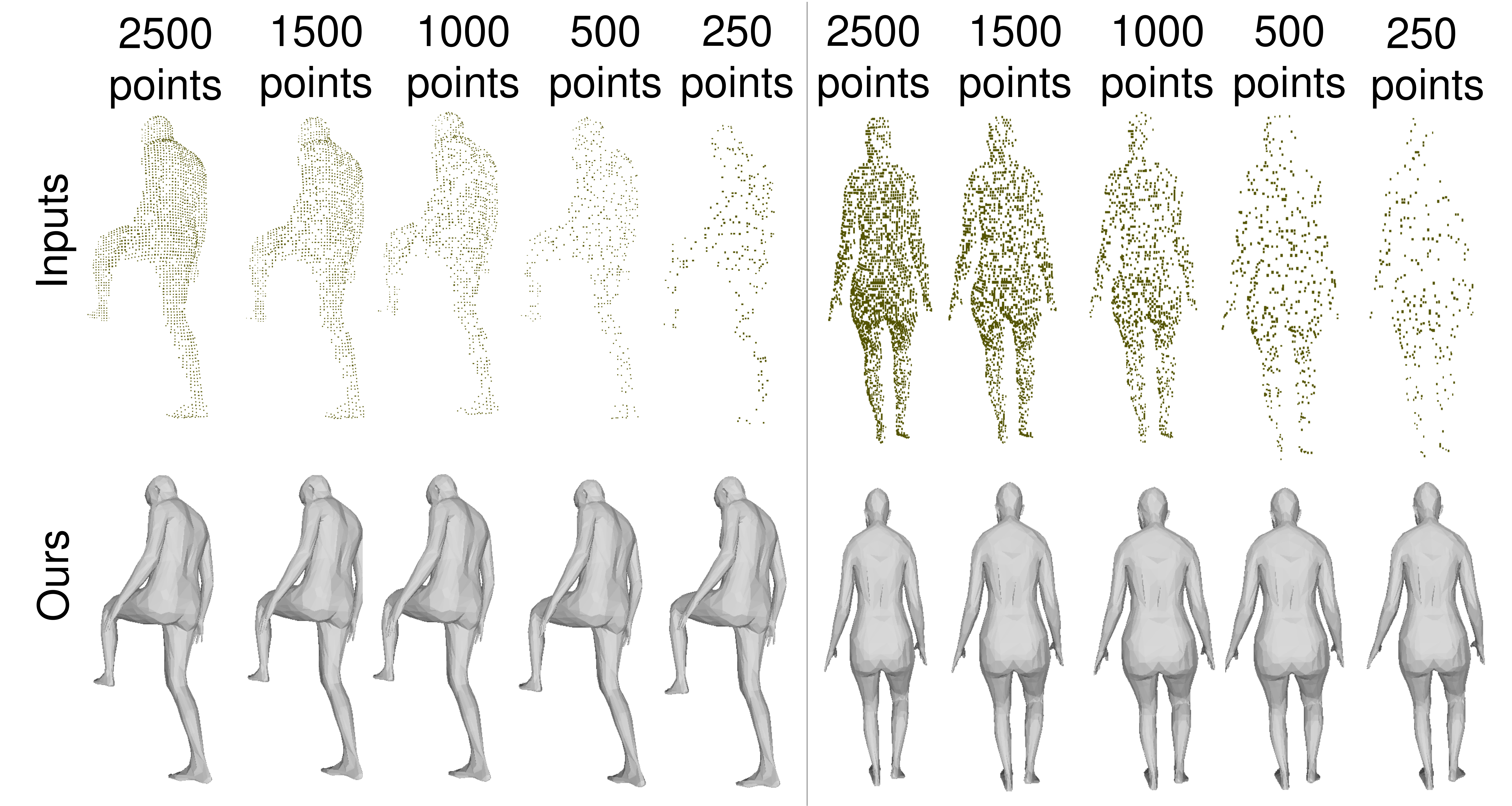}
\caption{Qualitative results by applying our VoteHMR to varying numbers of points. For each result, from left to right, we show the reconstruction results when the input point clouds have $2,500$, $1,500$, $1,000$, $500$, $250$ points, respectively.}
\label{size_visualization}
\end{figure}

\subsubsection{Reconstruction against Varying Point Numbers.}
\label{ssub:reconstruction_against_varying_point_numbers_}

Figure~\ref{noise_figure}~(right) shows the PVE scores of different point cloud based methods against the varying sizes of the input point clouds, on the SURREAL test set.
The input point clouds are randomly sampled to different numbers. And each method is trained when the input point cloud has $2,500$ points.
The comparison results demonstrate the robustness of our model even when only a fraction of points exists in the test set. The visualization results in Figure~\ref{size_visualization} also show that our model performs well when the input point clouds only have $250$ points.




\section{Conclusion}
In this paper, we address the problem of robust human mesh recovery from a single-frame partial point cloud, by a novel occlusion-aware voting network, named as \method, to handle the complex self-occlusion and noises of point cloud data captured from commodity-level depth sensors.
The experimental results have demonstrated that our method can achieve state-of-the-art performance on SURREAL and DFAUST datasets, and generalize well to real data captured by depth sensors like Berkeley MHAD dataset, without the aid of further weakly supervised fine-tuning. 

\subsection*{Acknowledgement}
This work was supported by the Key Research and Development Program of Guangdong Province, China (No. 2019B010154003), and the National Natural Science Foundation of China (No. 61906012).

\vfill\eject
\bibliographystyle{ACM-Reference-Format}
\balance
\bibliography{new_reference.bib}

\end{document}